\documentclass[sigconf]{acmart}
\AtBeginDocument{%
  }
\usepackage{multirow}
\usepackage{subcaption}
\acmConference[ISMSI '25]{April 26--28,
  2025}{Tokyo, Japan}

\renewcommand\footnotetextcopyrightpermission[1]{}

\begin{document}

\title{FGATT: A Robust Framework for Wireless Data Imputation Using Fuzzy Graph Attention Networks and Transformer Encoders}


\author{Jinming Xing}
\affiliation{%
  \institution{North Carolina State University}
  \country{}}
\email{jxing6@ncsu.edu}

\author{Chang Xue}
\affiliation{%
  \institution{Yeshiva University}
  \country{}}
\email{cxue@mail.yu.edu}

\author{Dongwen Luo}
\affiliation{%
  \institution{South China University of Technology}
  \country{}}
\email{976267567ldw@gmail.com}

\author{Ruilin Xing}
\affiliation{%
  \institution{Guangxi University}
  \country{}}
\email{ruilinxing8@gmail.com}

\renewcommand{\shortauthors}{Jinming et al.}

\begin{abstract}
  Missing data is a pervasive challenge in wireless networks and many other domains, often compromising the performance of machine learning and deep learning models. To address this, we propose a novel framework FGATT that combines the Fuzzy Graph Attention Network (FGAT) with the Transformer encoder to perform robust and accurate data imputation. FGAT leverages fuzzy rough sets and graph attention mechanisms to capture spatial dependencies dynamically, even in scenarios where predefined spatial information is unavailable. The Transformer encoder is employed to model temporal dependencies, utilizing its self-attention mechanism to focus on significant time-series patterns. A self-adaptive graph construction method is introduced to enable dynamic connectivity learning, ensuring the framework's applicability to a wide range of wireless datasets. Extensive experiments demonstrate that our approach outperforms state-of-the-art methods in imputation accuracy and robustness, particularly in scenarios with substantial missing data. The proposed model is well-suited for applications in wireless sensor networks and IoT environments, where data integrity is critical.
\end{abstract}



\maketitle

\section{Introduction}
Missing data is a widespread problem across various domains, ranging from wireless sensor networks to healthcare and finance. It often occurs due to hardware malfunctions, transmission errors, or environmental factors, leading to incomplete datasets. These missing values can severely degrade the performance of Machine Learning (ML) and Deep Learning (DL) models, as most algorithms require complete data for training and inference. Data imputation techniques have thus become essential to restore the integrity of datasets by estimating and filling in missing values during preprocessing. Existing data imputation approaches fall into two broad categories: traditional statistical methods and temporal deep learning-based methods. Statistical approaches, such as ARIMA \cite{aa14arima}, SVR \cite{SS16svr}, and LR \cite{dc12lr}, rely on explicit mathematical formulations and are suited for specific applications. In contrast, temporal DL-based methods, including FFN \cite{vas17attention}, LSTM \cite{yu19lstm}, GRU \cite{chung14gru}, and Transformer models \cite{amatriain23transformer}, utilize sequential dependencies in time-series data to predict missing values.

While statistical methods are computationally efficient and interpretable, they are often limited in their capacity to handle nonlinearities and large-scale datasets. Temporal DL models, on the other hand, have demonstrated significant success in capturing time-series patterns and modeling complex relationships. However, they tend to overlook spatial dependencies, which are crucial in many real-world applications. For instance, in wireless sensor networks, the relationship between sensors' locations significantly influences the quality and accuracy of data imputation. The inability to incorporate such spatial information reduces the effectiveness of these models when applied to spatially structured data.

Graph Neural Networks (GNNs) have emerged as a powerful tool to incorporate relational and spatial dependencies in data. Among these, Fuzzy Graph Attention Networks (FGAT) \cite{xing24enhancing} represent a novel advancement by integrating fuzzy rough sets with graph attention mechanisms to address uncertainties in spatial relationships. FGAT enhances node representation learning, enabling it to capture more robust and discriminative features. However, spatial information is often unavailable in practical scenarios, requiring the design of dynamic graph construction methods for adaptive connectivity learning. To handle temporal dependencies, the Transformer encoder is integrated into the architecture, leveraging its self-attention mechanism to focus on critical temporal features. This hybrid approach addresses the limitations of existing methods by combining spatial and temporal modeling capabilities.

The main contributions of this paper are summarized as follows:
\begin{itemize}
    \item \textbf{Dynamic Graph Construction:} A novel self-adaptive connectivity learning approach that eliminates the need for predefined spatial structures.
    \item \textbf{Hybrid Framework Design:} An innovative combination of FGAT for spatial dependencies and Transformer encoders for temporal dependencies.
    \item \textbf{Comprehensive Evaluation:} Extensive experiments demonstrating superior imputation accuracy and robustness compared to state-of-the-art methods.
\end{itemize}

\section{Related Work}
Statistical methods have been widely used for data imputation due to their simplicity and interpretability. Techniques such as ARIMA, SVR, and LR estimate missing values based on observed patterns in the data. ARIMA, for instance, models temporal dependencies through autoregressive and moving average components, making it effective for time-series data. Similarly, SVR and LR predict missing values using regression-based approaches. Despite their advantages, these methods are inherently limited to linear relationships and struggle with high-dimensional, nonlinear, or large-scale datasets. Furthermore, they often require strong assumptions about data distribution and stationarity, which may not hold in real-world applications.

Deep Learning models \cite{cheng24patch,cheng25unifying,cheng22deep,cheng22estimation,xing24traffic,cheng23shapnn,yangbotnet,linresearch,yangrac} have found innovative applications across various fields like image analysis \cite{zhaoresearch, yangcnn}, virtual reality \cite{yangzhan}, sequences modeling \cite{dengresearch}, medical diagnosis \cite{yang24tcell}, and emotion recognition \cite{yangarxiv1}. Specifically, temporal models like LSTM networks and GRU are designed to capture long-term dependencies in time-series data by maintaining a memory of past information. Transformers, with their self-attention mechanisms, further advance this capability by allowing the model to focus on the most relevant parts of the sequence, irrespective of their distance. These temporal models excel in handling sequential relationships but often ignore spatial dependencies inherent in many datasets. For instance, in wireless sensor networks, where spatial proximity affects the relationships between sensor readings, the lack of spatial awareness in temporal-only models limits their imputation accuracy and robustness. Additionally, their performance may degrade in scenarios with extensive missing data, where the absence of spatial context makes temporal modeling less effective.

Graph neural networks have recently emerged as a promising approach for data imputation by capturing spatial and relational dependencies. Spatio-Temporal Graph Neural Networks (ST-GNNs) extend GNN architectures to model both spatial and temporal dimensions of the data. These methods construct graphs based on predefined spatial relationships, such as sensor locations or physical proximity, and then apply temporal models like RNNs or attention mechanisms to handle time-series dependencies. While effective in many cases, ST-GNNs often rely heavily on prior knowledge of spatial structures, which may not always be available or accurate in real-world datasets. Moreover, their reliance on static graph structures makes them less adaptable to dynamic environments. To address these limitations, based on FGAT \cite{xing24enhancing}, the Fuzzy Graph Attention-Transformer Network (FGATT) is proposed. FGATT introduces fuzzy rough sets to account for uncertainties in spatial data and dynamically learns graph structures to enhance spatial representation learning. Despite these advancements, Transformer encoder is adopted for temporal dependencies modeling, achieving a holistic approach to spatial-temporal imputation.

\section{Methodology}
In this section, we describe the foundational concepts and design of our framework. First, the principles of Fuzzy Rough Sets (FRS) are outlined, which form the basis of the dynamic graph construction scheme. We then detail the architecture of the Fuzzy Graph Attention Network (FGAT). Finally, the Transformer encoder is introduced to model temporal dependencies effectively.

\subsection{Dynamic Graph Construction}
As stated in \cite{xing22weighted,gao22param}, the fuzzy lower and upper approximations of a sample $x$ with respect to a group of samples $d_i$ under the attribute set $B$ are defined as follows:
\begin{equation}
    \begin{split}
        &\underline{R_B}d_i(x) = \inf_{y\in U}\max(1 - R(x,y), d_i(y)),\\
        &\overline{R_B}d_i(x) = \sup_{y\in U}\min(R(x,y), d_i(y))
    \end{split}
\end{equation}
where $R$ is typically a kernel function \cite{xing22weighted}.

To capture dynamic spatial relationships, we compute connectivity scores between nodes at each timestep. Given two nodes $i, j$ and their embeddings $x_i^t, x_j^t \in \mathbb{R}^d$ at time $t$, the connectivity score is defined as follows \cite{xing24enhancing}:
\begin{equation}
    Score^t(i,j) = \alpha \times \underline{R_B}d_j(x_i^t) + (1 - \alpha) \times \underline{R_B}d_i(x_j^t)
\end{equation}
where $\alpha$ is a hyperparameter balancing the two components.

The generated graph must represent the connectivity relationships over a predefined temporal context window $T$. To achieve this, a pooling mechanism aggregates per-time connectivity information into a single representation. As discussed in \cite{xing24pooling}, the three main pooling mechanisms are:
\begin{itemize}
    \item \textbf{Mean pooling:} Averages token embeddings for balanced representation.
    \item \textbf{Max pooling:} Captures salient features by selecting maximum values across dimensions.
    \item \textbf{Weighted Sum pooling:} Learns weights dynamically but requires additional computational resources.
\end{itemize}

Following recommendations in \cite{xing24pooling}, we use Mean pooling for aggregation. Specifically, the aggregated connectivity score for nodes $i$ and $j$ is computed as:
\begin{equation}
    Score(i,j) = \frac{1}{T} \sum_{t=1}^T Score^t(i,j)
\end{equation}

To optimize graph construction, only the top $K$ edges based on connectivity scores are retained, as excessive edges can degrade graph convolution performance by introducing noise and increasing training costs \cite{xing24enhancing}. Additionally, self-loops are removed, as they may lead to error accumulation in certain cases \cite{xing24enhancing}. This ensures a clean, efficient graph structure for downstream processing.

\subsection{Fuzzy Graph Attention Network (FGAT)}
The Fuzzy Graph Attention Network (FGAT) layer, as proposed in \cite{xing24enhancing} and depicted in Figure \ref{fig:FGATT}, integrates multiple components to enable robust learning. These include:
\begin{figure}
    \centering
    \includegraphics[width=0.99\linewidth]{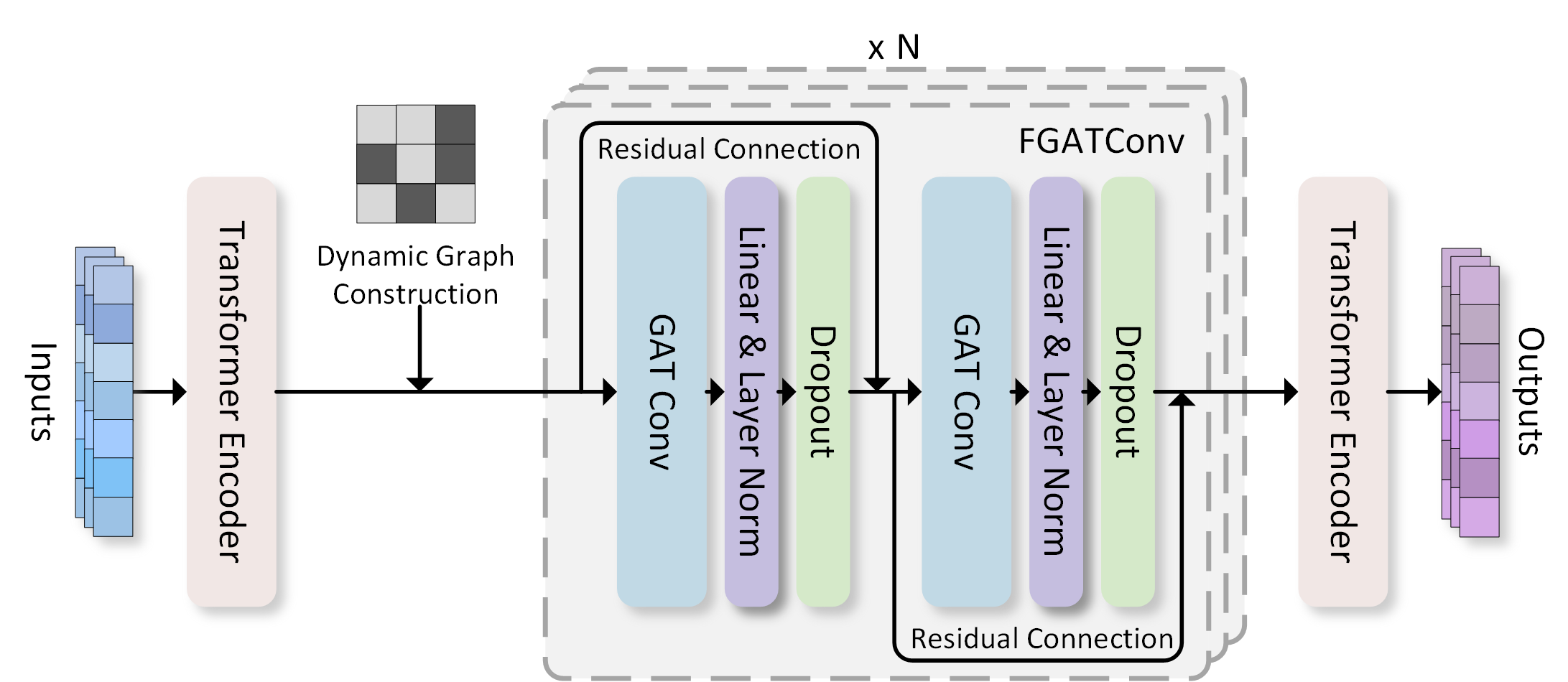}
    \caption{Fuzzy Graph Attention-Transformer Network}
    \label{fig:FGATT}
\end{figure}
\begin{itemize}
    \item \textbf{Graph Attention Network (GAT) layers:} Aggregate information from neighboring nodes while assigning learnable attention weights to prioritize important connections.
    \item \textbf{Linear and Normalization layers:} Standardize outputs and ensure stability during training.
    \item \textbf{Dropout layers:} Enhance generalization by mitigating overfitting.
    \item \textbf{Residual connections:} Facilitate gradient flow and improve model training efficiency.
\end{itemize}

The GAT mechanism is formally defined as:
\begin{equation}
    h_i^{'} = \text{LeakyReLU} \left( \sum_{j \in \mathcal{N}(i)} \alpha_{ij} W h_j \right)
\end{equation}
where $h_i^{'}$ is the updated embedding for node $i$, $\alpha_{ij}$ is the attention weight between nodes $i$ and $j$, $W$ is the learnable weight matrix, and $\mathcal{N}(i)$ denotes the set of neighbors for node $i$.

Layer normalization ensures numerical stability and faster convergence by normalizing input activations as follows:
\begin{equation}
    y = \frac{x - \mu}{\sqrt{\sigma^2 + \epsilon}} \cdot \gamma + \beta
\end{equation}
where $\mu$ and $\sigma$ are the mean and variance of the input $x$, and $\gamma$ and $\beta$ are learnable parameters.

\subsection{Transformer Encoder}
To effectively capture global and local temporal dependencies, the Transformer encoder is employed. Its self-attention mechanism allows the model to focus on relevant timesteps while aggregating bi-directional information efficiently. Compared to decoder-only or encoder-decoder architectures, the encoder-only structure achieves a balance between computational cost and information richness.

The self-attention mechanism is defined as:
\begin{equation}
    \text{Attention}(Q, K, V) = \text{softmax} \left( \frac{QK^\top}{\sqrt{d_k}} \right) V
\end{equation}
where $Q$, $K$, and $V$ are the query, key, and value matrices, and $d_k$ is the dimensionality of $K$. The mechanism ensures that the model attends to the most informative parts of the input sequence.

The Transformer encoder comprises multi-head attention layers followed by feedforward networks, which are both normalized and enhanced by residual connections. This architecture enables the model to process sequential data efficiently while preserving contextual information.

\section{Experiments}
In this section, we evaluate the performance of the proposed FGATT framework. First, we introduce the datasets used in our study, followed by a description of the experimental settings and evaluation metrics. Finally, we present and discuss the results.

\subsection{Datasets}
The Secure Water Treatment (SWaT) dataset \cite{ap16swat} is a widely recognized benchmark for data imputation and anomaly detection tasks. This dataset contains time-series data such as water levels, flow rates, and chemical concentrations collected by wireless sensors. Missing values are introduced due to various causes, including sensor malfunctions, network disruptions, and cyberattacks.
\begin{table}[htbp]
    \centering
    \caption{Dataset Summarization}
    \begin{tabular}{lccc}
        \toprule
        Dataset    & \#Samples & \#Nodes & Granularity \\
        \midrule
        SWaT.A7.22 & 3600      & 28      & 1 Sec       \\
        SWaT.A7.29 & 7201      & 25      & 1 Sec       \\
        \bottomrule
    \end{tabular}%
    \label{tab:dataset summarization}%
\end{table}%

For our experiments, two sub-datasets from SWaT are selected, each representing distinct operational settings. The details of these datasets are summarized in Table \ref{tab:dataset summarization}. These datasets provide a robust testbed to evaluate the effectiveness of our proposed method under realistic conditions of missing data.

\subsection{Experiment Settings and Evaluation Metrics}
To ensure fair evaluation and avoid dominance by particular sensor data, all datasets are preprocessed using min-max normalization. After normalization, the datasets are split into training, validation, and testing sets, comprising 70\%, 10\%, and 20\% of the data, respectively. The context length for each sample is set to 16 timesteps, providing sufficient temporal information for the models.

\begin{figure*}[htbp]
    \centering
    \begin{subfigure}[htbp]{0.32\linewidth}
        \centering
        \includegraphics[width=\linewidth]{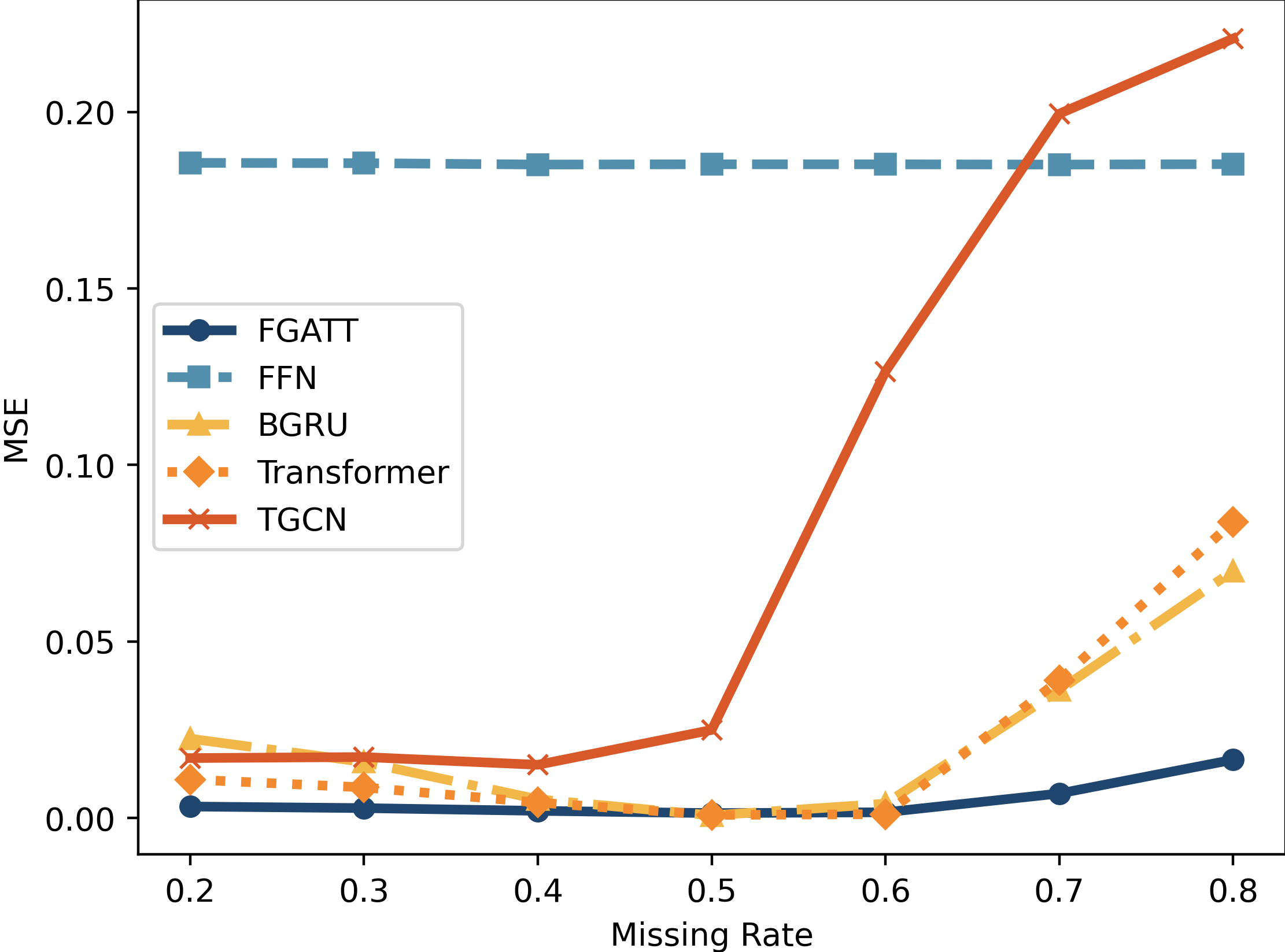}
        \caption{Mean Squared Error}
        \label{fig:22MSE}
    \end{subfigure}
    \hfill
    \begin{subfigure}[htbp]{0.32\linewidth}
        \centering
        \includegraphics[width=\linewidth]{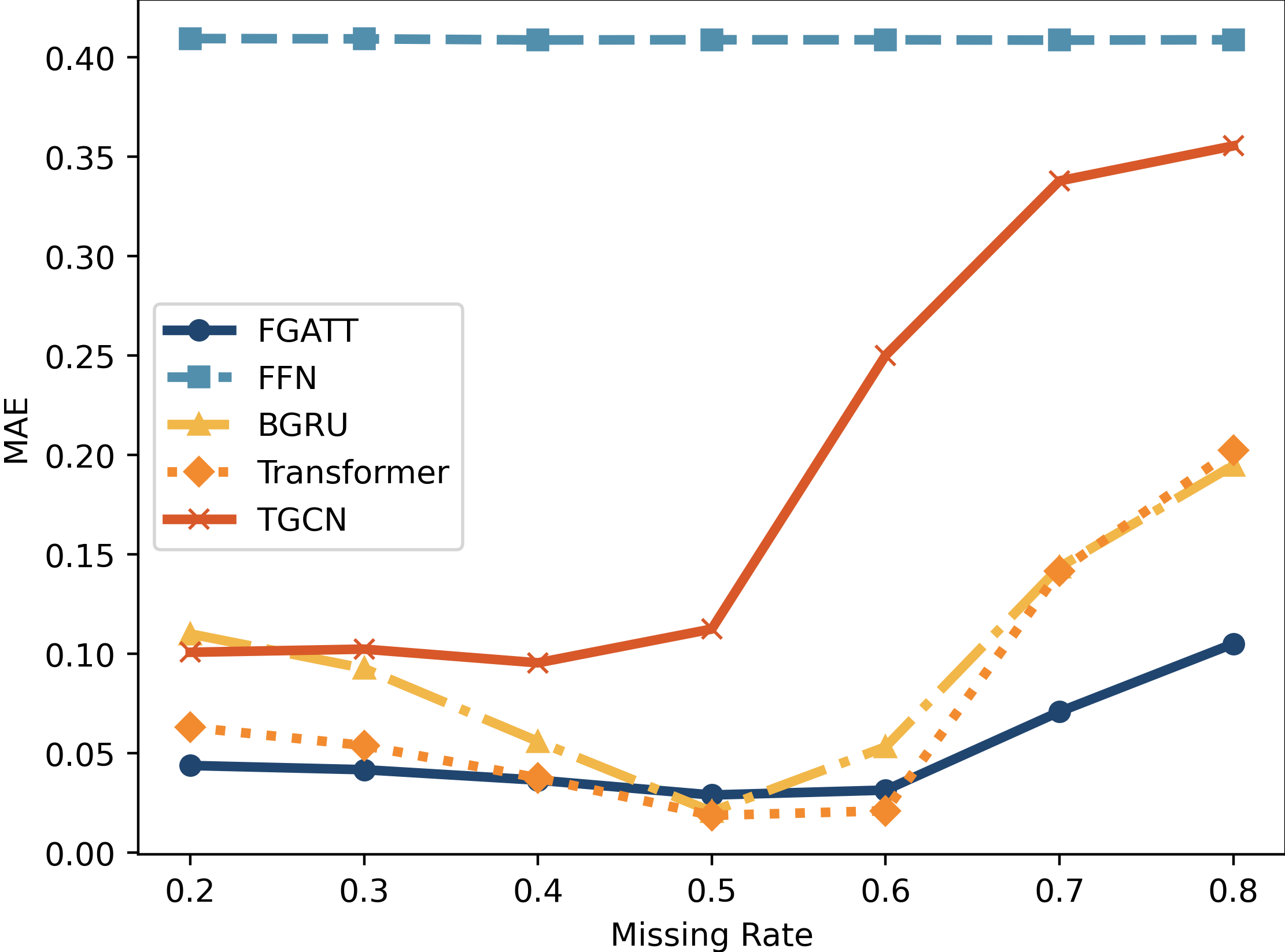}
        \caption{Mean Absolute Error}
        \label{fig:22MAE}
    \end{subfigure}
    \hfill
    \begin{subfigure}[htbp]{0.32\linewidth}
        \centering
        \includegraphics[width=\linewidth]{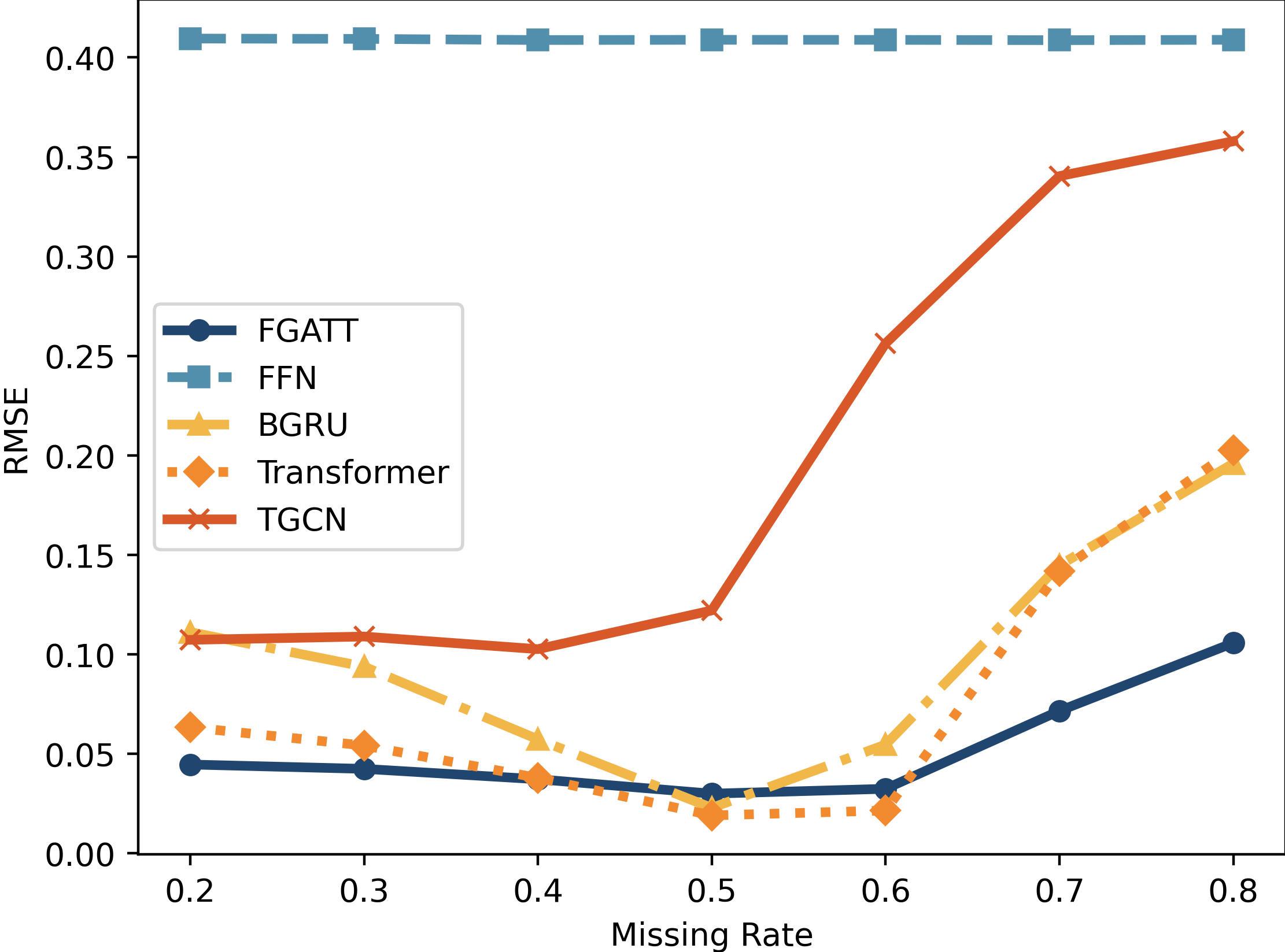}
        \caption{Root Mean Squared Error}
        \label{fig:22RMSE}
    \end{subfigure}
    \caption{Performance Evaluation on SWaT.A7.22}
    \label{fig:22}
\end{figure*}
\begin{figure*}[htbp]
    \centering
    \begin{subfigure}[htbp]{0.32\linewidth}
        \centering
        \includegraphics[width=\linewidth]{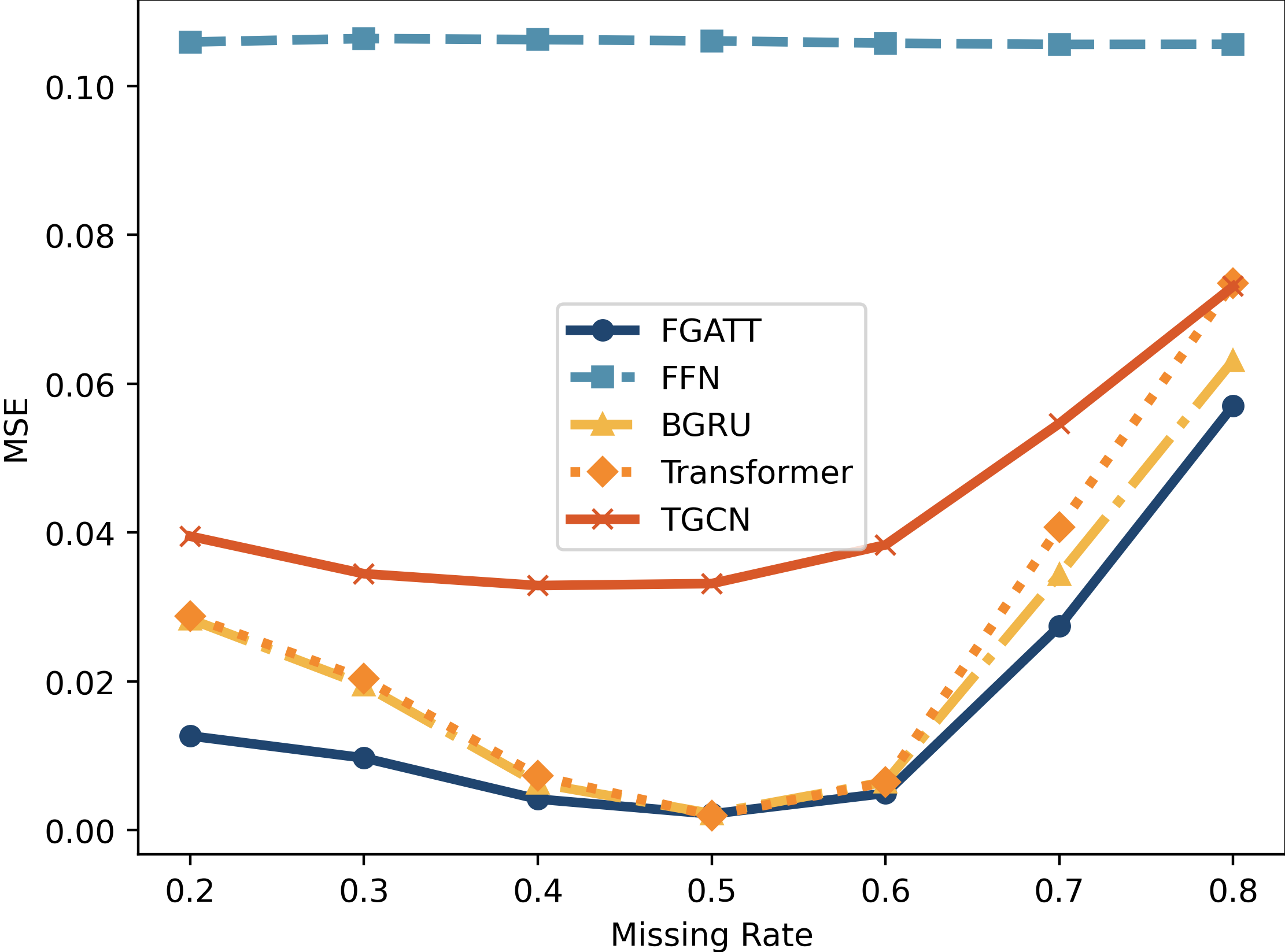}
        \caption{Mean Squared Error}
        \label{fig:29MSE}
    \end{subfigure}
    \hfill
    \begin{subfigure}[htbp]{0.32\linewidth}
        \centering
        \includegraphics[width=\linewidth]{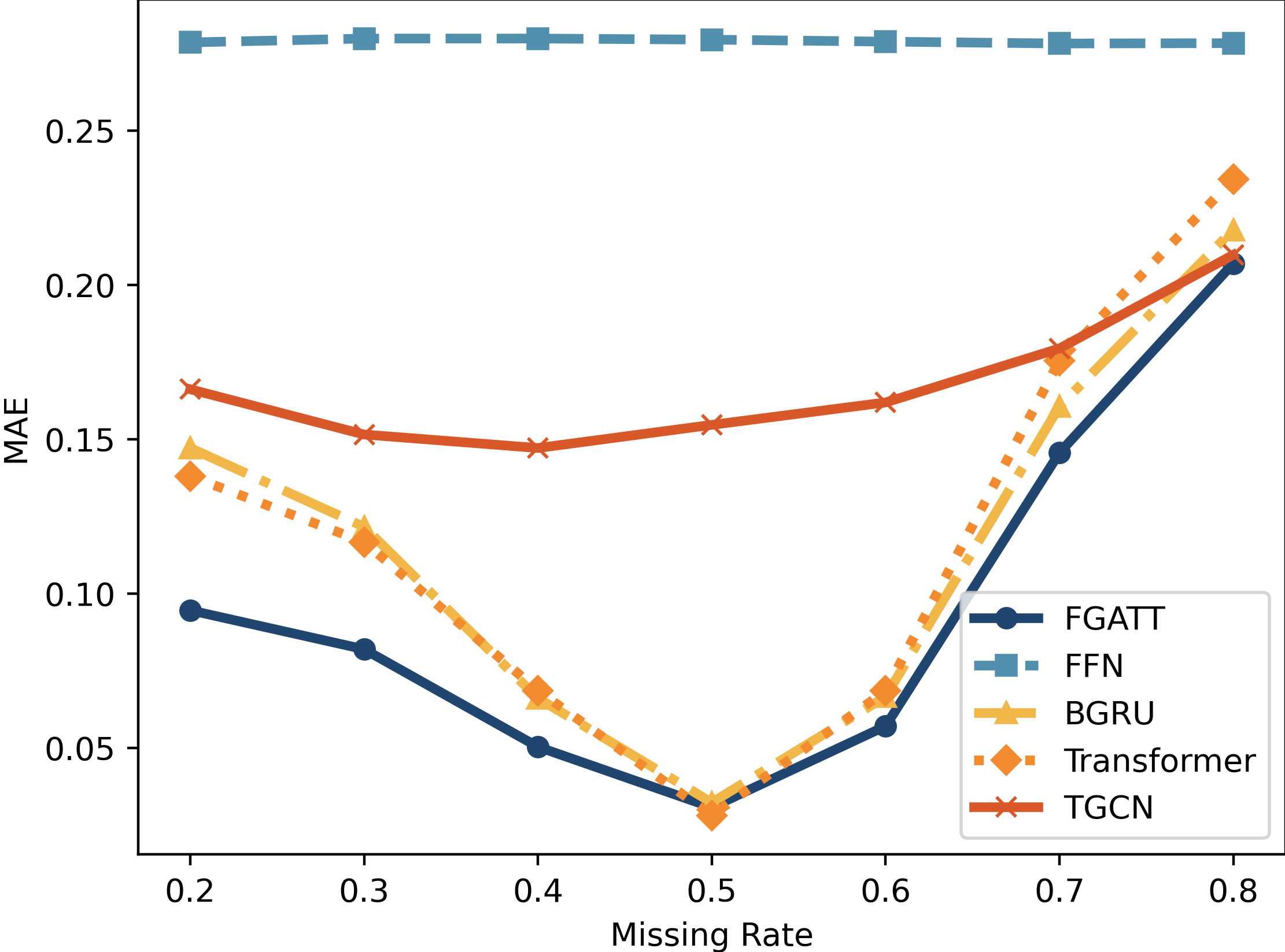}
        \caption{Mean Absolute Error}
        \label{fig:29MAE}
    \end{subfigure}
    \hfill
    \begin{subfigure}[htbp]{0.32\linewidth}
        \centering
        \includegraphics[width=\linewidth]{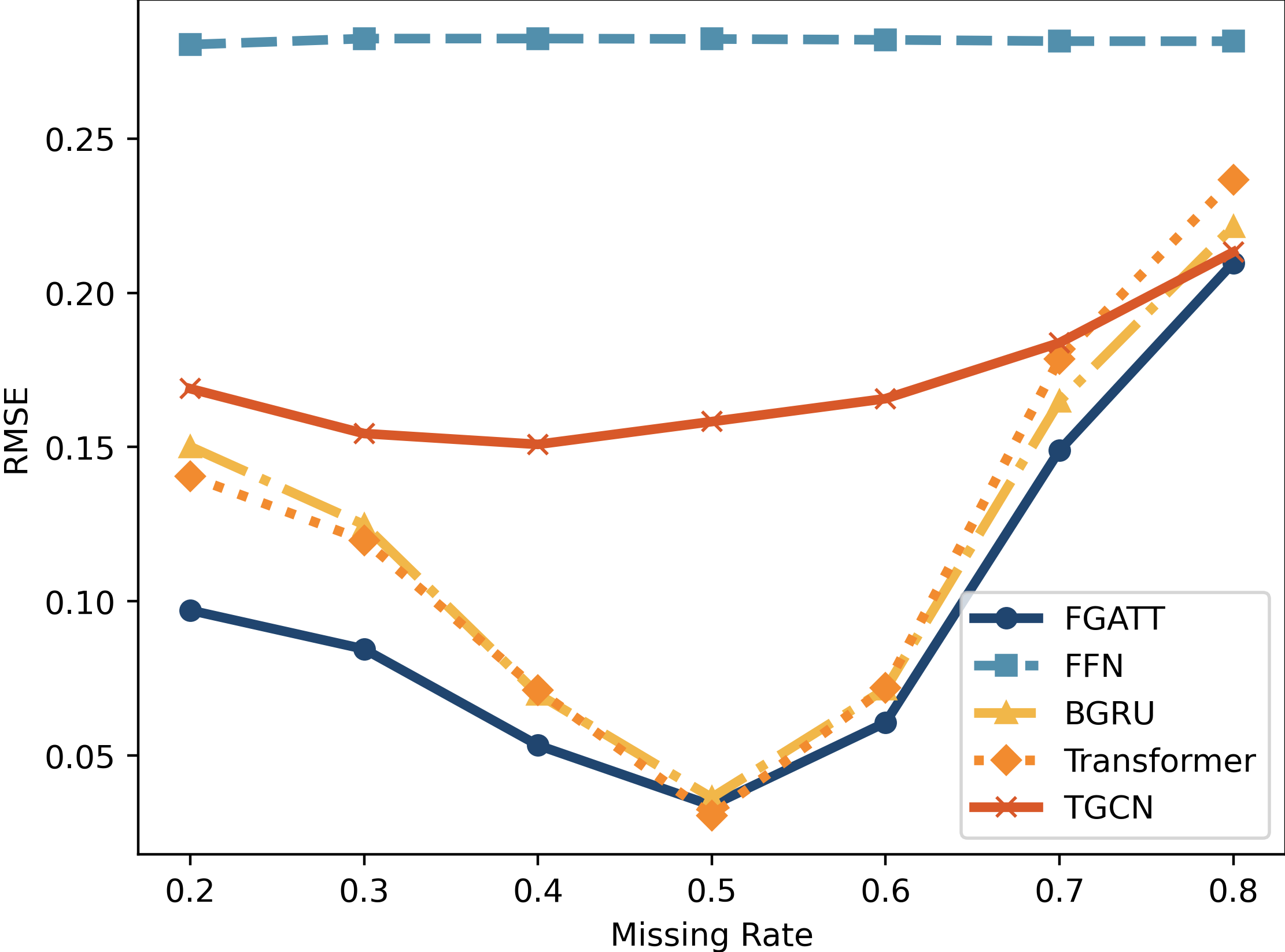}
        \caption{Root Mean Squared Error}
        \label{fig:29RMSE}
    \end{subfigure}
    \caption{Performance Evaluation on SWaT.A7.29}
    \label{fig:29}
\end{figure*}

To benchmark our method, we compare FGATT with four widely recognized baselines:
\begin{itemize}
    \item \textbf{FFN} \cite{vas17attention}: A feedforward network with simple architecture prone to overfitting.
    \item \textbf{BGRU} \cite{li22bigru}: A bidirectional GRU model capable of capturing temporal dependencies.
    \item \textbf{Transformer} \cite{amatriain23transformer}: A state-of-the-art temporal model leveraging self-attention mechanisms.
    \item \textbf{TGCN} \cite{zhao23tgcn}: A spatial-temporal GNN designed to incorporate spatial and temporal relationships.
\end{itemize}

During training, the missing rate is set to 50\%, simulating conditions where half of the data points are unavailable. For testing, each model is evaluated under varying missing rates, ranging from 20\% to 80\% in increments of 10\%.

To assess model performance comprehensively, we use three metrics:
\begin{itemize}
    \item \textbf{Mean Squared Error (MSE)}: Measures the average squared difference between predicted and actual values.
    \item \textbf{Mean Absolute Error (MAE)}: Captures the average absolute difference, offering an intuitive understanding of error magnitude.
    \item \textbf{Root Mean Squared Error (RMSE)}: Highlights the impact of larger errors by emphasizing their contribution.
\end{itemize}

\subsection{Results}
The comparison results for different missing rates are presented in Figures \ref{fig:22} and \ref{fig:29}. The following key observations are derived:

\textbf{Performance on A7.22 Dataset:}
FGATT demonstrates the best average performance across all metrics. For missing rates below 50\%, FGATT maintains stable errors, as it was trained with a 50\% missing rate. This robustness indicates the model's ability to generalize well under less challenging conditions. Other baselines, including BGRU, Transformer, and TGCN, exhibit similar patterns but with higher errors compared to FGATT. FFN, due to its simplistic architecture, fails to capture underlying patterns and exhibits poor performance across all scenarios.

As the missing rate exceeds 50\%, the performance of all methods degrades, which is expected as the available information becomes increasingly sparse. However, FGATT continues to outperform the baselines by leveraging its fuzzy rough sets-based dynamic graph and FGAT framework. Notably, TGCN experiences the steepest performance drop, reflecting its limited ability to adapt to dynamic and uncertain environments.

\textbf{Performance on A7.29 Dataset:}
On the A7.29 dataset, FGATT consistently achieves the best performance under all metrics, reaffirming its superiority. Interestingly, all models show a tendency to overfit at the 50\% missing rate, with performance worsening as the missing rate deviates in either direction. For missing rates below 50\%, as the rate decreases, models generally perform worse, likely due to overfitting to the 50\% missing rate during training. Similarly, performance declines for missing rates above 50\%, as the increasing sparsity of data reduces the amount of useful information.

Despite these trends, FGATT exhibits remarkable robustness, outperforming baselines by a significant margin. This can be attributed to its fuzzy rough sets-based dynamic graph construction, which captures meaningful spatial dependencies, and the FGAT framework, which efficiently aggregates spatial and temporal

\section{Conclusion}
This paper presents a robust and effective framework for wireless data imputation, combining the strengths of Fuzzy Graph Attention Networks and Transformer-based temporal modeling. By introducing a dynamic graph construction method, the model addresses the limitations of traditional spatial-temporal techniques, which often rely on predefined spatial structures. The integration of fuzzy rough sets with graph attention enhances spatial representation learning, while the Transformer encoder captures complex temporal dependencies. Experimental results validate the superiority of our approach, demonstrating its ability to achieve high imputation accuracy even under challenging conditions with extensive missing data. Future work will explore extending the framework to handle real-time imputation tasks and evaluating its performance in diverse IoT applications.



\begin{thebibliography}{00}
    \bibitem{aa14arima} Ariyo, Adebiyi A., Adewumi O. Adewumi, and Charles K. Ayo. "Stock price prediction using the ARIMA model." In 2014 UKSim-AMSS 16th international conference on computer modelling and simulation, pp. 106-112. IEEE, 2014.
    \bibitem{SS16svr} Suthaharan, Shan, and Shan Suthaharan. "Support vector machine." Machine learning models and algorithms for big data classification: thinking with examples for effective learning (2016): 207-235.
    \bibitem{dc12lr} Montgomery, Douglas C., Elizabeth A. Peck, and G. Geoffrey Vining. Introduction to linear regression analysis. John Wiley \& Sons, 2021.
    \bibitem{vas17attention} Vaswani, A. "Attention is all you need." Advances in Neural Information Processing Systems (2017).
    \bibitem{yu19lstm} Yu, Yong, Xiaosheng Si, Changhua Hu, and Jianxun Zhang. "A review of recurrent neural networks: LSTM cells and network architectures." Neural computation 31, no. 7 (2019): 1235-1270.
    \bibitem{chung14gru} Chung, Junyoung, Caglar Gulcehre, KyungHyun Cho, and Yoshua Bengio. "Empirical evaluation of gated recurrent neural networks on sequence modeling." arXiv preprint arXiv:1412.3555 (2014).
    \bibitem{amatriain23transformer} Amatriain, Xavier, Ananth Sankar, Jie Bing, Praveen Kumar Bodigutla, Timothy J. Hazen, and Michaeel Kazi. "Transformer models: an introduction and catalog." arXiv preprint arXiv:2302.07730 (2023).
    \bibitem{cheng24patch} Cheng, Qisen, Shuhui Qu, and Janghwan Lee. "Patch-aware Vector Quantized Codebook Learning for Unsupervised Visual Defect Detection." In *2024 IEEE 36th International Conference on Tools with Artificial Intelligence (ICTAI)*, pp. 586-592. IEEE, 2024.
    \bibitem{cheng25unifying} Cheng, Qisen, Jinming Xing, Chang Xue, and Xiaoran Yang. "Unifying Prediction and Explanation in Time-Series Transformers via Shapley-based Pretraining." *arXiv preprint arXiv:2501.15070* (2025).
    \bibitem{cheng22deep} Cheng, Qisen, Shuhui Qu, and Janghwan Lee. "72‐3: Deep Learning Based Visual Defect Detection in Noisy and Imbalanced Data." In *SID Symposium Digest of Technical Papers*, vol. 53, no. 1, pp. 971-974. 2022.
    \bibitem{cheng22estimation} Cheng, Qisen, Chang Zhang, and Xiang Shen. "Estimation of Energy and Time Usage in 3D Printing With Multimodal Neural Network." In *2022 4th International Conference on Frontiers Technology of Information and Computer (ICFTIC)*, pp. 900-903. IEEE, 2022.
    \bibitem{xing24traffic} Xing, Jinming, Zhaomin Xiao, Yingyi Wu, Jinran Zhang, Zhuoer Xu, and Zhelu Mai. "Network Traffic Forecasting via Fuzzy Spatial-Temporal Fusion Graph Neural Networks." In *2024 11th International Conference on Soft Computing \& Machine Intelligence (ISCMI)*, pp. 282-286. IEEE, 2024.
    \bibitem{cheng23shapnn} Cheng, Qisen, Shuhui Qu, and Janghwan Lee. "SHAPNN: Shapley Value Regularized Tabular Neural Network." *arXiv preprint arXiv:2309.08799* (2023).
    \bibitem{yangbotnet}Yang, Xiaoran, Zhen Guo, and Zetian Mai. "Botnet detection based on machine learning." In 2022 International Conference on Blockchain Technology and Information Security (ICBCTIS), pp. 213-217. IEEE, 2022.
    \bibitem{linresearch}Lin, Chengrong, Shaofan Chen, Xiaoran Yang, Caimao Li, Cong Qu, and Qiuhong Chen. "Research and application of knowledge graph technology for intelligent question answering." In 2021 12th International Symposium on Parallel Architectures, Algorithms and Programming (PAAP), pp. 152-156. IEEE, 2021.
    \bibitem{yangrac}Yang, Xiaoran. "Research on automatic composition based on multiple machine learning models." In 2021 3rd International Conference on Artificial Intelligence and Advanced Manufacture, pp. 1206-1209. 2021.
    \bibitem{zhaoresearch}Zhao, Sihao, Jiahui Xie, Chu Lin, Xiaolan Nie, Jun Ye, Xiaoran Yang, and Pengzhi Xu. "Image Retrieval Based on Blockchain." In 2022 International Conference on Blockchain Technology and Information Security (ICBCTIS), pp. 210-212. IEEE, 2022.
    \bibitem{yangcnn}Liu, Juncheng, Xiaoran Yang, Jing-Chen Hong, and Hiroyasu Iwata. "Analysis of Lifting Posture by Two Inertial Measurement Units and a Classification Model Based on a Convolutional Neural Network." In 2024 10th IEEE RAS/EMBS International Conference for Biomedical Robotics and Biomechatronics (BioRob), pp. 383-388. IEEE, 2024.
    \bibitem{yangzhan}Yang, Xiaoran, Yang Zhan, Yukiko Iwasaki, Miaohui Shi, Shijie Tang, and Hiroyasu Iwata. "Balancing Real-world Interaction and VR Immersion with AI Vision Robotic Arm." In 2023 IEEE International Conference on Mechatronics and Automation (ICMA), pp. 2051-2057. IEEE, 2023.
    \bibitem{dengresearch}Deng, Xiaoru, Hui Zhou, Xiaoran Yang, and Chunyang Ye. "Short-Term Traffic Condition Prediction Based on Multi-source Data Fusion." In International Conference on Data Mining and Big Data, pp. 327-335. Singapore: Springer Singapore, 2021.
    \bibitem{yang24tcell} Yang, Jing‐Min, Nan Zhang, Tao Luo, Mei Yang, Wen‐Kang Shen, Zhen‐Lin Tan, Yun Xia et al. "TCellSI: A novel method for T cell state assessment and its applications in immune environment prediction." *Imeta* 3, no. 5 (2024): e231.
    \bibitem{yangarxiv1}Yang, Xiaoran, Shuhan Yu, and Wenxi Xu. "Improvement and Implementation of a Speech Emotion Recognition Model Based on Dual-Layer LSTM." arXiv preprint arXiv:2411.09189 (2024).
    \bibitem{xing24enhancing} Xing, Jinming. "Enhancing Link Prediction with Fuzzy Graph Attention Networks and Dynamic Negative Sampling." arXiv preprint arXiv:2411.07482 (2024).
    \bibitem{gao22param} Gao, Can, et al. "Parameterized maximum-entropy-based three-way approximate attribute reduction." International Journal of Approximate Reasoning 151 (2022): 85-100.
    \bibitem{xing22weighted} Xing, Jinming, Can Gao, and Jie Zhou. "Weighted fuzzy rough sets-based tri-training and its application to medical diagnosis." Applied Soft Computing 124 (2022): 109025.
    \bibitem{xing24pooling} Xing, Jinming, Ruilin Xing, and Yan Sun. "Comparative Analysis of Pooling Mechanisms in LLMs: A Sentiment Analysis Perspective." arXiv preprint arXiv:2411.14654 (2024).
    \bibitem{ap16swat} Mathur, Aditya P., and Nils Ole Tippenhauer. "SWaT: A water treatment testbed for research and training on ICS security." In 2016 international workshop on cyber-physical systems for smart water networks (CySWater), pp. 31-36. IEEE, 2016.
    \bibitem{li22bigru} Li, Xuechen, Xinfang Ma, Fengchao Xiao, Cong Xiao, Fei Wang, and Shicheng Zhang. "Time-series production forecasting method based on the integration of Bidirectional Gated Recurrent Unit (Bi-GRU) network and Sparrow Search Algorithm (SSA)." Journal of Petroleum Science and Engineering 208 (2022): 109309.
    \bibitem{zhao23tgcn} Zhao, Ling, Yujiao Song, Chao Zhang, Yu Liu, Pu Wang, Tao Lin, Min Deng, and Haifeng Li. "T-GCN: A temporal graph convolutional network for traffic prediction." IEEE transactions on intelligent transportation systems 21, no. 9 (2019): 3848-3858.
\end{thebibliography}
\end{document}